%% file: conference_101719.tex
\documentclass[letterpaper, 10 pt, conference]{ieeeconf}
\IEEEoverridecommandlockouts
\usepackage[T1]{fontenc}
\usepackage[utf8]{inputenc}
\usepackage{svg}

\usepackage{amsmath,amssymb,amsfonts}

\usepackage[ruled,vlined,linesnumbered]{algorithm2e}
\DontPrintSemicolon

\usepackage{xcolor}
\usepackage{comment}
\usepackage{tikz}
\usetikzlibrary{calc,patterns,angles,quotes,3d}
\usepackage{tkz-euclide}
\usepackage{tikz-3dplot}
\usepackage{booktabs}
\usepackage{graphicx}
\usepackage{subcaption}

\usepackage[backend=bibtex,citestyle=numeric-comp,bibstyle=ieee,sorting=none,minbibnames=6,maxbibnames=6,defernumbers=true,giveninits=true,doi=false,isbn=false,url=false,eprint=false]{biblatex}
\usepackage[hidelinks]{hyperref}

\DeclareUnicodeCharacter{FFFC}{}

\newcommand\gderror[1]{
   \typeout{--------------------------------------------------------------------}
   \typeout{------- #1 ---------}
   \typeout{--------------------------------------------------------------------}
   {\bf #1}
}
\newcounter{gdTmp} 
\newcounter{gdLastCount}
\newcommand\maxpage[2][Error]{  %  arg1: error message, arg2: max page number,
\ifnum\value{page}>#2
    \gderror{On page {\thepage} we are past page #2 (too long).   #1 }
\else\fi
\setcounter{gdLastCount}{\value{page}} % used to count skipped articles
}
\newcommand\maxpageSinceLast[2][Error]{  %  arg1: error message, arg2: max page number,
\ifnum \numexpr \value{page} - \value{gdLastCount}\relax>#2
    \gderror{Exceeds max length #2 pages. Page \thepage: #1}
\thepage\else\fi
\setcounter{gdLastCount}{\value{page}} % used to count skipped articles
}

\def\BibTeX{{\rm B\kern-.05em{\sc i\kern-.025em b}\kern-.08em
    T\kern-.1667em\lower.7ex\hbox{E}\kern-.125emX}}

\begin{document}

% \makeatletter
% \newcommand{\newlineauthors}{%
%   \end{@IEEEauthorhalign}\hfill\mbox{}\par
%   \mbox{}\hfill\begin{@IEEEauthorhalign}
% }
% \makeatother

\title{Stable Multi-Drone GNSS Tracking System for Marine Robots}

\author{
Shuo Wen$^{1, 2 \dag *}$  \quad
Edwin Meriaux$^{1,2, 3 \dag *}$ \quad
Mariana Sosa Guzmán$^{1, 2}$*\\
Zhizun Wang$^{1, 2}$\quad
Junming Shi$^{1, 2}$\quad
Gregory Dudek$^{1, 2}$%
\thanks{\noindent  $\dag$ Corresponding authors * Co-first authors.  $^{1}$ McGill University $^{2}$ MILA-Quebec AI Institute $^{3}$ Université Paris-Saclay, CentraleSupélec. Emails: shuo.wen@mail.mcgill.ca; edwin.meriaux@mail.mcgill.ca}
}

\maketitle
\begin{abstract}
% Accurate localization is crucial for marine robotics, yet traditional onboard Global Navigation Satellite System (GNSS) approaches are difficult or ineffective due to signal reflection on the water’s surface and its high cost of aquatic GNSS receivers. Existing approaches, such as inertial navigation, Doppler Velocity Loggers (DVL), SLAM, and acoustic-based methods, face challenges like error accumulation and high computational complexity. Therefore, a more efficient and scalable solution remains necessary. 
%we develop and evaluate an approach to pose estimate based on using aerial drones to estimate the positions on vehicles below them.  We focus particularly on vehicles that are on the ocean surface, or below it. 

%.

%Accurate localization is critical for marine robotics, but traditional Global Navigation Satellite System (GNSS) methods are often unreliable or inconvenient at the surface of the water and unmanageable in submerged tasks. Alternative approaches such as inertial navigation, Doppler Velocity Loggers (DVL), SLAM, and acoustic systems suffer from error accumulation or high complexity. These trade-offs motivate the search for approaches that balance reliability, cost, and scalability. This paper proposes an alternative approach that leverages aerial drones equipped with GNSS localization to track and localize marine robots once it is near the surface of the water. Our results show that this novel adaptation enables accurate single and multi-robot marine robot localization.  

Stable and accurate tracking is essential for marine robotics, yet Global Navigation Satellite System (GNSS) signals vanish immediately below the sea surface. Traditional alternatives suffer from error accumulation, high computational demands, or infrastructure dependence. In this work, we present a multi-drone GNSS-based tracking system for surface and near-surface marine robots. Our approach combines efficient visual detection, lightweight multi-object tracking, GNSS-based triangulation, and a confidence-weighted Extended Kalman Filter (EKF) to provide stable GNSS estimation in real time. We further introduce a cross-drone tracking ID alignment algorithm that enforces global consistency across views, enabling robust multi-robot tracking with cooperative aerial coverage. We validate our system in diversified complex settings to show the accuracy and robustness of the proposed algorithm.

% \greg{Add a tiny bit more details (like one sentence or two), see review comment.}
\end{abstract}

\section{Introduction}

While satellite-based positioning is widely accepted for surface marine robots, its effectiveness diminishes once the robots descend even a very short distance below the ocean surface or if the antenna is wet with salt water. This fundamental limitation prevents marine robots from maintaining continuous GNSS coverage, creating a critical gap in localization capability. As a result, robust alternatives are essential to ensure accurate and persistent positioning in real-world marine environments. 

A variety of onboard localization methods have been proposed, including inertial sensors combined with DVLs \cite{turbulent_ocean,artic_dvl}. SLAM-based methods have been applied to improve autonomy, but are computationally expensive and unstable in dynamic environments \cite{slam_fail1,slam_fail2}. Acoustic techniques such as Long Baseline (LBL) and Ultra Short Baseline (USBL) localization rely on fixed beacons \cite{lbl3,lbl2}. While effective under certain conditions, these methods suffer from issues such as accumulated drift, reliance on infrastructure, and high computational cost.
% An extensive body of work has explored onboard localization for marine robots. Inertial sensors with Doppler Velocity Loggers (DVLs) are widely used but accumulate drift over time \cite{turbulent_ocean}, \cite{artic_dvl}. Acoustic methods such as Long Baseline (LBL) and Ultra Short Baseline (USBL) rely on fixed beacons \cite{lbl3}, \cite{lbl2}, while SLAM-based approaches improve autonomy but face high computational cost and reduced robustness in dynamic environments \cite{slam_fail1}, \cite{slam_fail2}.

% To overcome these limitations, offboard localization, where sensing and computation are shifted to external platforms offers a promising solution. This strategy reduces the cost and power burden on the robot and opens the door for aerial–underwater cooperation, such as drone-mounted GNSS support for recalibration. 

\begin{figure}[t]
    \centering
    \includegraphics[width=0.48\textwidth, height=0.19\textheight]{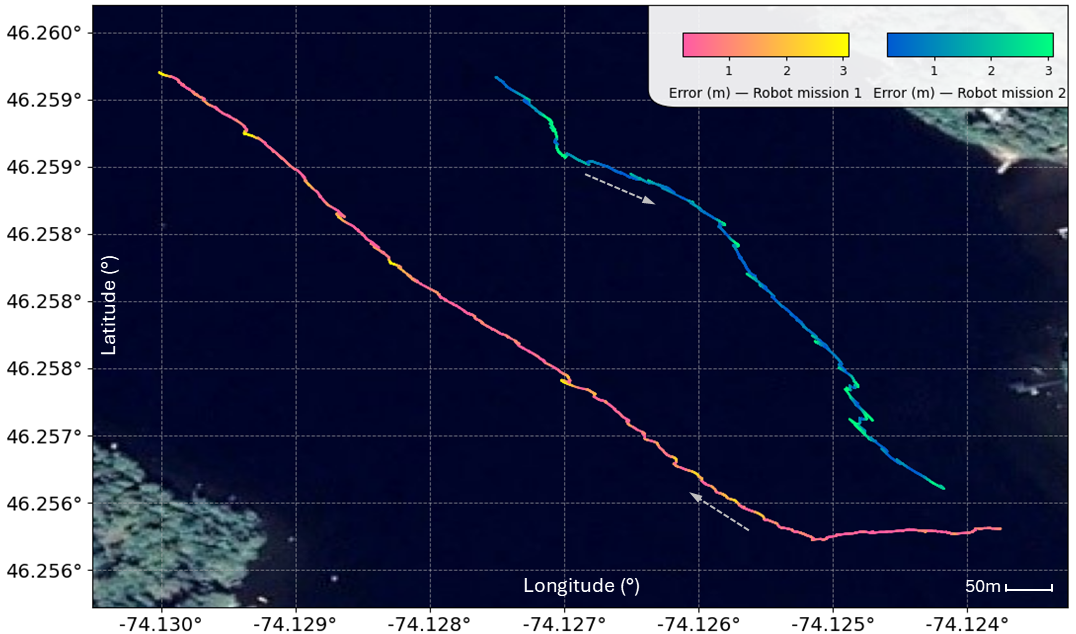} % Adjust width as needed
    \caption{GNSS trajectory estimates of the same marine robot across two trials from Categories 1 and 2 in open-water bodies. The color gradient represents localization error, with brighter colors indicating larger errors. White arrows mark the direction of motion along each trajectory. The mean error is 0.842 m for Mission 1 and 1.282 m for Mission 2.}
    \label{money}
\end{figure}

To mitigate these drawbacks, offboard localization transfers sensing and computation to external platforms, reducing the hardware and energy burden on the underwater robot. This opens the possibility for aerial–underwater cooperation, with drones serving as mobile positioning beacons. Prior work has demonstrated the feasibility of using a single drone for this purpose \cite{wen2025scalable}, but relying on a single viewpoint makes the system susceptible to occlusions, limited coverage, and frequent tracking loss, while also creating a single point of failure. Our system instead employs multiple drones, ensuring overlapping coverage and redundancy: if one drone temporarily loses sight of the target, the others can continue providing accurate observations.

\begin{figure}[h]
    \centering
    \includegraphics[width=0.48\textwidth, height=0.2\textheight]{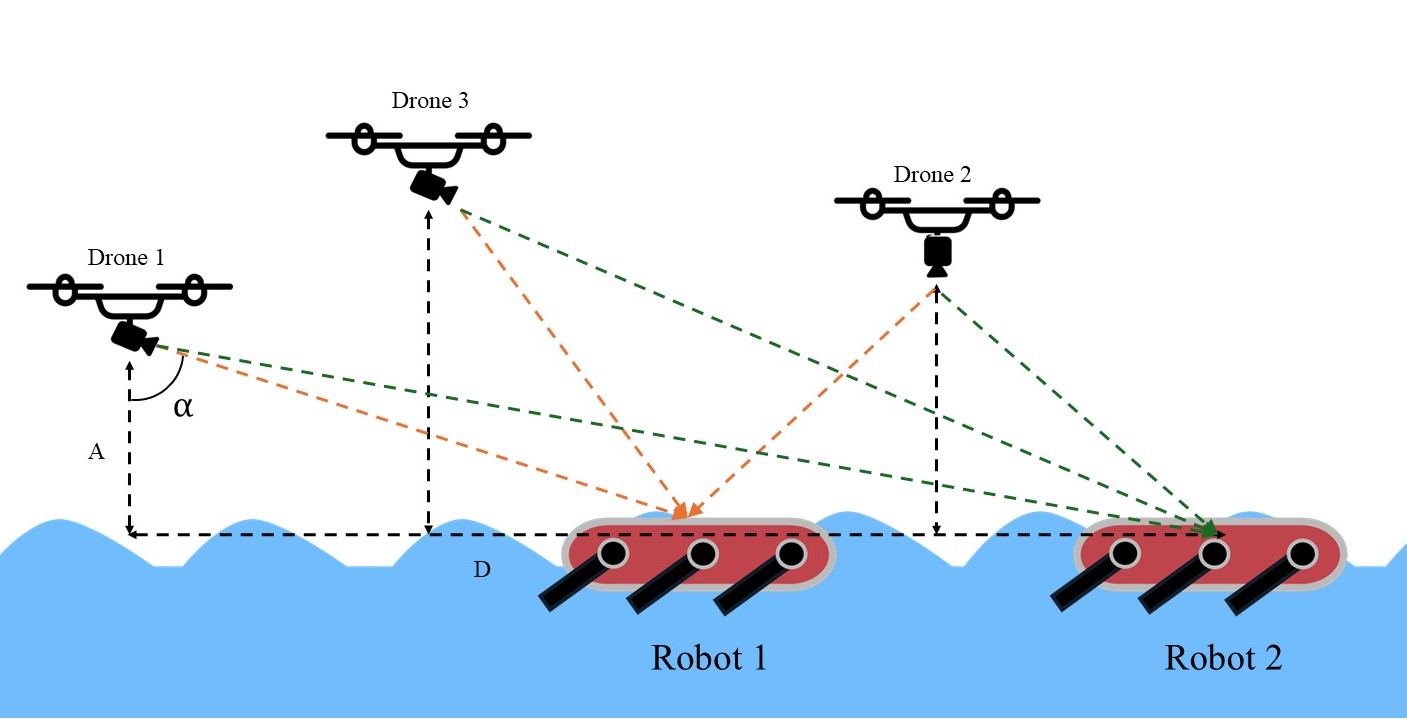} % Adjust width as needed
    \caption{Tracking system with three aerial drones collaboratively localizing two marine robots operating near the water surface.}
    \label{cartoon}
\end{figure}

In this paper, we describe and evaluate a system for estimating the GNSS position of mobile vehicles, focused specifically on marine robots, as long as they remain close enough to the surface to remain visually observable. Building on prior work done in aerial localization \cite{wen2025scalable}, we extend and adapt it to a multi-drone, multi-robot setting, providing a robust, scalable, and low-cost solution for surface-level marine tracking. An illustration of the setup is shown in Fig.~\ref{cartoon}. To ensure stable ID assignment in long-range tracking tasks, we introduce a hybrid matching strategy that combines weighted similarity scores from both image and GNSS domains. We further develop an ID alignment algorithm to synchronize robot tracking IDs across different drones. Finally, by integrating the Extended Kalman Filter with confidence-based estimation aggregation, our system achieves an average relative localization error of less than 1 m in the most ideal trials, and only 1.7 m in the more challenging cases. Sample estimated data can be seen in Fig.~\ref{money}, where 2 sample tracking missions are shown. The framework applies to both submerged and surface vehicles, as well as non-instrumented objects such as humans in search and rescue scenarios or marine objects. Experiments are conducted using two marine robots~\cite{dudek2007aqua}
from Independent Robotics, and the implementation code can be found at \url{https://github.com/stevvwen/stable_multidrone\ } with the data found at~\cite{meriaux2026crv}.

\section{Background}
%% motivation
Accurate localization of marine robots, marine creatures, and even people in a Human Robot Interaction (HRI) contexts is crucial for many applications, including ocean exploration \cite{ocean_exploration}, environmental monitoring \cite{env_surveillance}, underwater infrastructure inspection \cite{inf_inspect}, marine research \cite{marine_research}, and search and rescue operations~\cite{multi_robots,divertrack}. 

% title subject of change
\subsection{Localization for Marine Robotics}
% Marine robotic localization has traditionally focused on enabling a robot to estimate its own position in underwater environments. Earlier approaches have relied on onboard sensing, balancing trade-offs between cost, complexity, robustness, and accuracy. A common approach combines inertial sensors with Doppler Velocity Loggers (DVLs), which estimate velocity relative to water but accumulate drift over time \cite{turbulent_ocean}, \cite{artic_dvl}, \cite{dvl_error}. To mitigate this, acoustic methods such as Long Baseline (LBL) and Ultra Short Baseline (USBL) rely on triangulating signals from beacons or transponders \cite{lbl3}, \cite{lbl2}, \cite{lbl1}. Extensions include model-aided localization for gliders, using kinematic models and single-beacon timing within an EKF to reduce infrastructure needs \cite{glider_model}. Newer frameworks, like 3D-BLUE \cite{3d_blue} and low-power 3D localization systems \cite{low_power_3d}, further improve scalability and efficiency using piezo-electric backscatter technology.
Marine robotic localization has traditionally focused on enabling a robot to estimate its own position underwater. Earlier approaches relied on onboard sensing, balancing trade-offs between cost, complexity, robustness, and accuracy. A common method combines inertial sensors with DVLs, which estimate velocity relative to water but accumulate drift over time \cite{turbulent_ocean,artic_dvl,dvl_error}. To mitigate this, acoustic methods such as Long Baseline (LBL) and Ultra Short Baseline (USBL) triangulate signals from beacons or transponders \cite{lbl3,lbl2,lbl1}. Extensions include model-aided localization for gliders, using kinematic models and single-beacon timing within an EKF \cite{glider_model}, while newer frameworks like 3D-BLUE \cite{3d_blue} and low-power 3D localization systems \cite{low_power_3d} improve scalability and efficiency via piezo-electric backscatter.

To reduce dependence on external infrastructure, Simultaneous Localization and Mapping (SLAM) has been explored with methods such as EKF, SEIF, and FastSLAM \cite{Salvi2008VisualSLAM,seif,fast}. A notable example is Sonar Visual Inertial SLAM, fusing stereo vision, IMU, and sonar for GNSS-free mapping of underwater structures \cite{sonar_vio}. While SLAM enhances autonomy, challenges remain in computation and robustness at large-scale or highly dynamic environments ~\cite{slam_fail1,slam_fail2,manderson2018vision}.

Beyond onboard approaches, hybrid strategies have also emerged, including cross-domain localization aligning visual data~\cite{boldo2024real,lu2025ctrnet}, whether it is aerial or water-based~\cite {cross_view}, cooperative bearing-only localization integrating inertial and depth sensors \cite{bearings}, and multi-robot frameworks where agents share mapping and exploration tasks to improve efficiency~\cite{multi_robots}.

Offboard methods shift sensing and computation to external systems, making them cheaper, lighter, and less power-demanding for the robot. A representative example is the use of aerial platforms. To mitigate drift, underwater robots need to surface periodically for positional recalibration, for which inexpensive drone-mounted GNSS modules offer a practical solution. UAV-based localization has been demonstrated with onboard object detection \cite{drone_loc} and cooperative systems like Sunflower, which combines aerial laser sensing with underwater communication \cite{sunflower}. While accurate and robust, Sunflower requires specialized laser infrastructure, which limits its scalability in resource-constrained settings.

% Offboard localization of a marine robot exists in one of two settings. The former is the one where the robot is informed of its position and the latter is when the robot is not informed. Informing the robot has the same difficulties as a submerged robot trying to receive GPS signal. The latter case—where the surface drone localizes the robot without transmitting its position back—is particularly useful in shallow-water scenarios such as aquaculture cage inspections, flood-response debris mapping, and diver guidance in ports and canals, where precise localization within 2 m at depths up to 3 m is sufficient for effective operations.
Offboard localization generally occurs in two modes: the robot may be informed of its estimated position, or remain uninformed while an external system tracks it. The latter is the focus of the paper and is particularly useful in shallow-water scenarios such as aquaculture inspections~\cite{hogue2006underwater}, flood responses, and more generally the tracking of near-surface objects.

%, where sub-meter precision is unnecessary and ~2 m accuracy at depths up to 3 m is sufficient.

\subsection{Multi Camera Multi Object Tracking (MCMOT)}
% Multi-Camera Multi-Object Tracking (MCMOT) addresses key limitations of single-camera systems, such as occlusions and restricted fields of view, by leveraging distributed cameras for broader coverage and more reliable localization. This has become increasingly important in surveillance, robotics, and autonomous systems.

% Early approaches relied on centralized fusion, exploiting 3D geometry for accurate tracking \cite{cai_mcmot} but suffering from high communication overhead and poor scalability. Distributed methods emerged to overcome these issues, including overlapping camera networks for activity monitoring \cite{collins_mcmot}.

% Research has since explored diverse fusion strategies: early feature fusion for crowded scenes \cite{bird-eye}, geometry-based triangulation for efficient 3D drone tracking \cite{Rosner2025}, and distributed schemes such as consensus-based estimation \cite{CVPR_2013}, decentralized inference \cite{casao}, and real-time association frameworks \cite{DMMA}, each balancing robustness, scalability, and efficiency. A notable line of work is decentralized tracking, where cameras share only compact state information rather than raw data \cite{taj_decentralized}, avoiding communication bottlenecks and enabling real-time scalability.

% Building on these insights, our work adopts a distributed MCMOT framework for multi-drone object localization, leveraging geometric consistency across views to achieve robust and scalable tracking in outdoor environments.

MCMOT overcomes the limitations of single-camera systems, such as occlusions and restricted views, by combining observations from distributed cameras for broader coverage and more reliable localization. Early work used centralized fusion of multi-camera streams \cite{cai_mcmot}, but scalability and latency issues led to distributed alternatives, including overlapping camera networks for activity monitoring \cite{collins_mcmot}. Subsequent research explored diverse fusion strategies, from feature fusion for crowded scenes \cite{bird-eye} and geometry-based triangulation for 3D drone tracking \cite{Rosner2025}, to consensus-based estimation \cite{CVPR_2013}, decentralized inference \cite{casao}, and real-time association frameworks \cite{DMMA}, each trading off robustness, scalability, and efficiency. A particularly relevant direction is decentralized tracking, where cameras share only compact state information rather than raw data \cite{taj_decentralized}, removing communication bottlenecks and enabling real-time scalability. Building on these insights, our work adopts a distributed MCMOT framework for multi-drone object localization, leveraging geometric consistency across views to achieve robust and scalable tracking in marine environments.

\subsection{Tracking While Moving}
% Tracking while moving has gained increasing attention in recent years, as both cameras and platforms (e.g., vehicles, drones, or robots) often operate under motion rather than static settings. This setting introduces additional challenges such as ego-motion, changing viewpoints, and frame misalignment, which complicate data association and identity consistency. In autonomous driving \cite{mcm3D} addresses these issues by introducing a Global Association Graph Model that jointly links detections across multiple moving cameras, reducing fragmented tracklets and refining 3D detection. Similarly, \cite{twemcm} tackles identity switches in moving-camera scenarios with Linker, a lightweight association model, and a color-transfer module for appearance consistency across cameras. In aerial tracking \cite{multiUAVs} proposes a fusion mechanism that adaptively weighs appearance, geometric, and distribution features to robustly associate targets across high-altitude UAV views. More broadly, collaborative robot systems such as \cite{peterson2025tcaff} focus on aligning local frames through temporal consistency, enabling distributed multi-object tracking among mobile robots even without prior knowledge of relative poses.
Tracking under motion has attracted increasing attention as cameras and platforms (e.g., vehicles, drones, robots) are now frequently deployed in dynamic rather than static settings. This shift introduces challenges such as ego-motion, viewpoint changes, and frame misalignment. In autonomous driving, significant effort has been devoted to mitigating identity switches through techniques such as Global Association Graphs~\cite{mcm3D} and linker models~\cite{zhang2023towards}. In the aerial domain, target association across multiple UAV views has been achieved by fusing appearance, geometric, and distribution features, while collaborative systems leverage temporal consistency to align robot frames without prior pose information~\cite{multiUAVs,peterson2025tcaff}. The marine environment presents additional difficulties due to complex image formation effects, further complicating robust tracking~\cite{manderson2018vision}.

\section{Methodology}
The proposed method to observe and estimate the location of the marine robots consists of three main stages: Data Acquisition, Visual Localization, and Estimation.

\subsection{Data Acquisition and Preprocessing}

We conducted freshwater field trials with our marine robots to collect data for training the vision model and validating our methods. A simple GNSS receiver, mounted at the robot’s center, recorded ground-truth signals. Meanwhile, three drones flew at different altitudes, capturing videos from multiple angles. The experiment setup is illustrated in Fig.~\ref{cartoon}. To test robustness across applications, we performed trials under varying experimental conditions.

We use established data augmentation techniques from prior work \cite{wen2025scalable} to enhance the robustness of our vision module training. We use generic augmentations (rotation, flipping, cropping, padding, and affine transformations) together with domain-specific ones: motion blur for fast movement of the drones, glass blur for water distortion, and brightness or color shifts for illumination variation. To maximize diversity, all augmentations are applied independently with assigned probabilities and executed in random order.

\subsection{Marine Robot Detection and Tracking}
% khanam2024yolov11
For the detection backbone, we fine-tune YOLO v-11~\cite{khanam2024yolov11} on the marine robot videos we collected. It is selected for its balance of accuracy, response speed, and lightweight design. This stepwise approach enables the model to build upon its initial knowledge, improving its ability to handle challenging underwater scenarios. The vision model is trained on a desktop equipped with an NVIDIA RTX 4090 GPU and an AMD Ryzen 9 9950X processor using the Ultralytics framework with default training and model hyperparameters. Input images are downsampled to 640×640 pixels to match the native training resolution of YOLO.

\begin{figure}[h]
    \centering
    \includegraphics[width=0.40\textwidth]{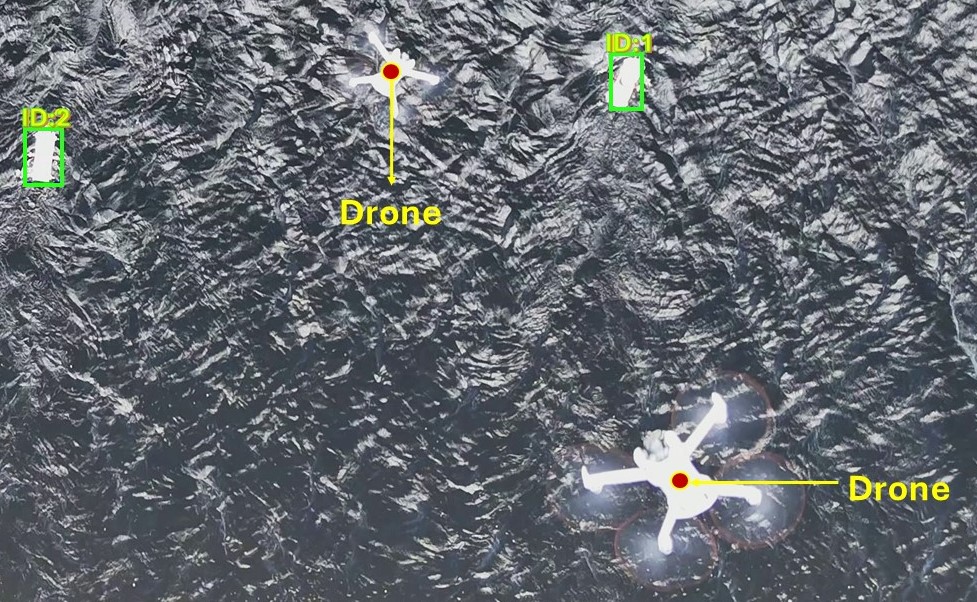} % Adjust width as needed
    \caption{Sample frame from the drone at the highest altitude, showing two tracked marine robots (ID:1 and ID:2) along with two drones at lower altitudes in view.}
    \label{sampleframe}
\end{figure}

We integrate the trained vision module into ByteTrack \cite{zhang2022bytetrack}, which we employ as our tracking module. ByteTrack distinguishes itself from other algorithms through its simplicity and computational efficiency, while maintaining reliable data association under uncertainty. This makes it particularly suitable for our marine robotics tracking tasks. Each drone runs ByteTrack independently, with one drone's detections seen in Fig.~\ref{sampleframe}. ID alignment is applied at the initial detection (see \ref{subsec:idcon}). The detection score threshold is set to $\tau = 0.5$, with a lower bound of 0.1 for the second association.

\subsection{GNSS Position Estimation}

To estimate GNSS coordinates, we follow the computational approach of \cite{wen2025scalable}. For an image of dimensions $W \times H$, where $W$ and $H$ denote width and height in pixels, the pixel coordinates $(x_o, y_o)$ of the object center are first obtained by vision models such as YOLO as described earlier. For simplicity, we assume the camera has zero yaw and roll angles.

The first step is to compute the object’s pixel displacement from the image center:
\begin{equation}
\Delta x = x_o - \frac{W}{2}, \quad \Delta y = y_o - \frac{H}{2}.
\end{equation}

Using the camera focal length $f$ and the sensor dimensions $(S_x, S_y)$, the object’s angular displacement from the image center in the $x$ and $y$ directions can be expressed as

\begin{equation}
\theta_x = \tan^{-1}\!\left(\frac{\Delta x}{W}\cdot \frac{S_x}{f}\right), \quad 
\theta_y = \tan^{-1}\!\left(\frac{\Delta y}{H}\cdot \frac{S_y}{f}\right)
\end{equation}

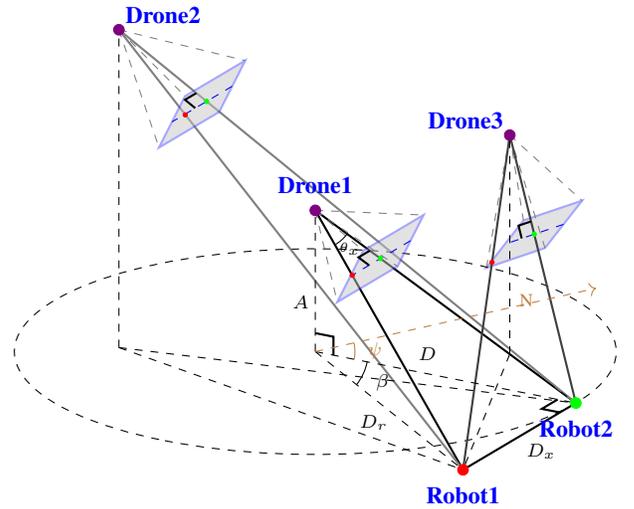
\begin{figure}[htbp]
  \centering
  \input{drawing.tex}
  \caption{Position estimation geometry. Image planes and projection rays illustrate the cameras' fields of view and triangulation. For simplicity, Robot 2 is centered in all views.} %, without affecting the mathematical formulation
  \label{fig:my-drawing}
\end{figure}

With the angular displacement between the drone to the object, the ground distance \(D\) to the object is
\begin{equation}
    D = A \tan(\alpha + \theta_y).
\end{equation}
where \(A\) is the drone altitude and $\alpha$ is the effective camera tilt relative to the vertical, combined with pitch.

The real-world distance corresponding to $\theta_x$ is given by:
\begin{equation}
D_x = \sqrt{A^2+D^2}\tan(\theta_x).
\end{equation}

% \frac{\sqrt{\sin^2(\theta_x)\cdot A^2+  D^2}}{\cos(\theta_x)}

The robot’s angular displacement relative to the drone and the radial ground distance are given as:

\begin{equation}
    \beta= \tan^{-1} ( \frac{D_x}{D}),  \qquad D_r = \sqrt{D^2+D_x^2}
\end{equation}

Fig.~\ref{fig:my-drawing} provides a 3D illustration of the computation on Drone 1 and Robot 1, augmented with additional information to depict the full geometry with multi-drone system.

The object’s displacement is then resolved into the global GNSS frame, where the drone’s heading $\psi$ relative to true north and $\beta$ yield the northward and eastward offset $\Delta N, \Delta E$:
\begin{equation}
\Delta N = D_r \cos(\psi + \beta), \qquad
\Delta E = D_r \sin(\psi + \beta)
\end{equation}

The estimated GNSS latitude and longitude coordinates are then obtained by directly incorporating the conversion from linear offsets to angular shifts:
\begin{equation}
\phi^{\text{est}} = \phi^{\text{drone}} + \frac{\Delta N}{\text{FPD}_{\text{lat}}}, \qquad
\lambda^{\text{est}} = \lambda^ {\text{drone}} + \frac{\Delta E}{\text{FPD}_{\text{lon}}}
\end{equation}
where $\text{FPD}_{\text{lat}}$ and $\text{FPD}_{\text{lon}}$ denote the feet-per-degree conversion factors for latitude and longitude, respectively.  

Our formulation provides a direct method for estimating a marine robot’s geolocation from aerial detections, based on the drone’s navigation and imaging parameters.

\subsection{Kalman Filter with Estimation Aggregation}
\label{subsec:kalmanfusion}

Tracking underwater robots requires combining noisy measurements from multiple sensing platforms into a consistent and robust state estimate. The Kalman filter provides a principled framework for this task, as it recursively fuses new positional measurements with prior state estimates while accounting for uncertainty~\cite{allotta2016unscented}. At each time step, multiple drones may detect the same marine robot, producing complementary position estimates. To consolidate these measurements, we compute a weighted average of the reported positions, where the weights are given by the normalized YOLO confidence scores of each detection. Formally, given GNSS coordinate estimates
$e_i= (\phi^{\text{est}}_i,\lambda^{\text{est}}_i)$ from each drone $i$ with detection confidences $c_i$, the fused estimate $e_{\text{final}}$ is defined as: $e_{\text{final}} = \frac{\sum_i c_i \, e_i}{\sum_i c_i}.$

We then pass $e_{\text{final}}$ through the aforementioned Kalman filter to get a stable filtered estimation. While the standard Kalman filter assumes linear system and measurement models, underwater dynamics and sensing modalities are often nonlinear (e.g., bearing and range observations, drag-dependent motion). To address these nonlinearities, we employ an Extended Kalman Filter (EKF)~\cite{ribeiro2004kalman}, which linearizes the motion and observation models around the current estimate at each step. This extension yields more accurate and stable tracking performance in our marine environment, ensuring that fused multi-drone observations are effectively integrated into the state estimation process. We implement an EKF with a discrete time step of 
$\Delta t = 0.1\mathrm{s}$, using a process noise acceleration variance of 
$q_{\mathrm{acc}} = 1.0$ to model system dynamics uncertainty, and a measurement 
noise variance of $r_{\mathrm{meas}} = 3.0$ to account for sensor observation errors.

\subsection{Minimize Tracking ID Switches with Hybrid Matching}
\label{subsec:hybrid}
A core step in ByteTrack is computing the similarity score for object matching, where we follow the original design and use Intersection Over Union (IOU) as the primary metric \cite{zhang2022bytetrack}. We omit Re-ID feature, which was discussed in ByteTrack~\cite{zhang2022bytetrack}, since it is computationally expensive for edge devices and ineffective for our robots, which share almost identical appearances. While the IOU is efficient and simple, camera shake caused by air turbulence and the rapid motions of drones and swimming robots can heavily affect its performance. A missed detection in a single frame can cause serious ID switching. To improve consistency, we fuse IOU scores with normalized GNSS distance scores. Formally, for each tracked object $t \in \mathcal{T}$ and detection $d \in \mathcal{D}$, we compute the pairwise IOU score $I(t, d)$ in the image space. We then combine this score with the GNSS-space distance $L(t, d)$, where the filtered GNSS position of $t$ is used as input instead of the raw estimation for stability. The final hybrid matching score is $s(t, d) = w_1 I(t, d) + w_2 L(t, d)$. We select the basic normalized Haversine distance \cite{chopde2013landmark} to be the distance function $L$. We choose $w_1= 0.7, w_2= 0.3$ to balance the reliability of IOU in the image space against the noise inherent in the GNSS space. 

%The distance function $L$ can be defined using either the normalized Haversine distance or the normalized Mahalanobis distance.

\begin{algorithm}[h]
\caption{ID alignment algorithm}
\label{alg:align}
\KwIn{Initialized tracks $\mathcal T$ from the reference drone, thresholds $\mu, \gamma$, drones $\{i\}$ with robot detections}
\KwOut{$\mathcal{Z}$: Set of tracking IDs for the drones}

$\mathcal{Z}= \{\}$

\For{\text{each drone i}}{
$\mathcal{D}= [d_j \in \textit{drone i}.\text{detections}]$

$M= [0]^{|\mathcal{D}| \times |\mathcal{T}|}$

\For{$d_j \in \mathcal D$}{
\For{$t_k \in \mathcal T$}{
    $\mathbf{M}_{j,k}= L(t_k, d_j)$\
}
}

$\text{ID}_{\text{matched}},\;\text{ID}_{\text{unmatched}}$= LinearAssignment($M, \mu$)

$\text{ID}_{\text{new}} = \text{Initialize}\_\text{ID}(\text{ID}_{\text{unmatched}}, \gamma)$

$\mathcal{Z}.\text{append}( \text{ID}_{\text{matched}} \cup \text{ID}_{\text{new}})$

}
\end{algorithm}

\subsection{Tracking with ID Consistency}
\label{subsec:idcon}
Each drone runs ByteTrack independently while sharing a unified ID counting system. This design ensures that objects maintain consistent identities across views, but without proper alignment, the same robot could be assigned different IDs in different drone frames. To avoid this mismatch, we designate one drone as the reference and initialize its tracks $t_i$. For every other drone, we compute the pairwise GNSS distance $L(t_i, d_j)$ between the reference tracks and its detections $d_j$. The Hungarian method \cite{kuhn1955hungarian} is then applied to solve the assignment problem and enforce consistent ID mapping across all drones, subject to a maximum matching threshold $\mu$. Unmatched detections are assigned new IDs only if their confidence score $c_j$ exceeds a threshold $\gamma$. The full procedure is summarized in Algorithm~\ref{alg:align}. In the experiments, we set $\mu= 30, \gamma= 0.6$.

\section{Results}

\subsection{Test Methodology}

We designed five categories of experiments to evaluate the accuracy and robustness of our tracking system under progressively more challenging conditions. \textbf{Category 1:} Two marine robots traversed long, mostly straight trajectories (up to 600m), while three drones tracked them overhead. These trials established a baseline with minimal changes in speed or heading. \textbf{Category 2:} The long-distance experiment was repeated with only two drones to assess the impact of reduced aerial coverage. \textbf{Category 3:} The experiment was repeated again with a single drone tracking both marine robots, providing insight into single-observer performance. \textbf{Category 4:} Two marine robots followed non-linear trajectories with three drones overhead, testing system performance under less structured motion. \textbf{Category 5:} A single marine robot executed a shorter trajectory with multiple sharp U-turns, challenging the system’s ability to handle rapid changes in both speed and direction. The results of these experiments can be seen in Table~\ref{tab:five_by_three}. Each test category is marked with the average error and standard deviation. 

\begin{table}[h]
  \centering
  \caption{Results Across All Test Categories}
  \label{tab:five_by_three}
  \begin{tabular}{lccc}
    \toprule
    \textbf{Test Categories} &\textbf{\# Drones} & \textbf{Avg. Error} & \textbf{Error STD} \\
    \midrule
    Linear Test 1 & 3 & 0.942 & 0.743 \\
    Linear Test 2 & 2 & 1.042 & 0.845 \\
    Linear Test 3 & 1 & 1.111 & 0.845 \\
    Non-Linear & 3 & 1.524 & 1.435 \\
    Hard Turns & 3 & 1.732 & 1.481 \\
    \bottomrule
  \end{tabular}
\end{table}

% \begin{figure}[t]
%     \centering
%     % \includegraphics[width=0.45\textwidth]{image_folder/Aqua.PNG} % Adjust width as needed
%     \includegraphics[width=0.5\textwidth, \textheight]{zoom2.png} % Adjust width as needed
%     \caption{The estimated trajectory translation from ICP to account for GNSS differences with the ground truth receivers.}
%     \label{icp}
% \end{figure}

\subsection{Error Estimation}

\begin{figure}[h]
    \centering
    \includegraphics[width=0.45\textwidth]{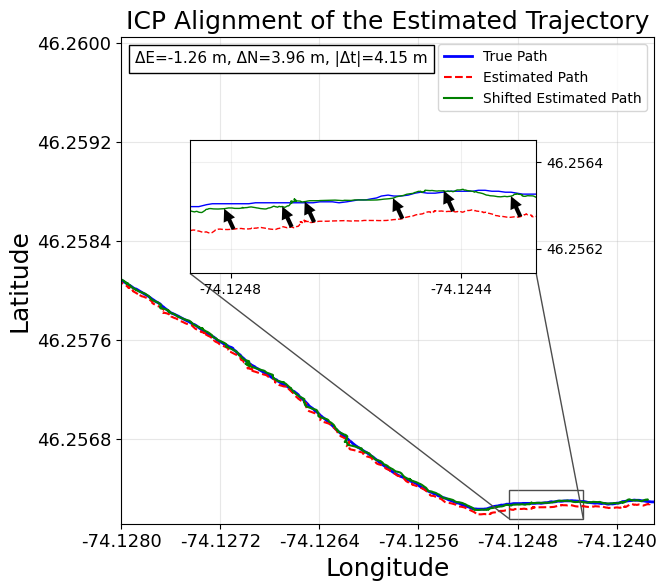} % Adjust width as needed
    \caption{Category 1 example showing ground truth and ICP-corrected trajectories, with black arrows indicating GNSS error corrections.}
    \label{icp}
\end{figure}

The GNSS estimation accuracy metric is defined as the spatial deviation between the true and estimated trajectories, both consisting of discrete data points sampled at different rates (10~Hz on the drones and 1~Hz on the GNSS receiver). Computing this error is not as simple as taking the haversine distance~\cite{chopde2013landmark} between points with matching timestamps, as (i) the clocks on the drones and GNSS receiver are not perfectly synchronized, with GNSS receiver timestamps delayed by up to 0.02~s, and (ii) a systematic GNSS offset existed between the two systems due to differences in receivers, satellite constellations, and error models. We use a translation-only variant of the Iterative Closest Point (ICP) algorithm~\cite{121791} to align two GNSS lines as seen in Fig.~\ref{icp} where the black arrows indicate the shift. Both lines are first converted from latitude/longitude into a common local East–North coordinate frame using an equirectangular projection around their mean location. At each iteration, every point in the estimation line is paired with its nearest neighbour in the true line using a KD-tree. The best translation for the current matches is then computed as the difference between the centroids of the matched point sets, and the estimated line is shifted accordingly. This process is repeated until the translation update becomes very small or a maximum number of iterations is reached, yielding aligned estimated GNSS values to properly compute the error between the two lines.

% \begin{figure}[h]
%     \centering
%     % \includegraphics[width=0.45\textwidth]{image_folder/Aqua.PNG} % Adjust width as needed
%     \includegraphics[width=0.45\textwidth]{zoom3.png} % Adjust width as needed
%     \caption{Category 1 example showing ground truth and ICP-corrected trajectories, with black arrows indicating GNSS error corrections.}
%     \label{icp}
% \end{figure}

\vspace{-10pt}

%To address these issues, we first applied Iterative Closest Point (ICP) to translate the estimated trajectory and minimize its alignment error with the true trajectory. Then, for each point on the estimated trajectory, we projected a perpendicular onto the \textit{piecewise-linear} true trajectory, treating consecutive true points as being connected by straight segments, and found the nearest point on this line (which may lie between two true data points). The residual haversine distances after ICP alignment were taken as the final error measure, providing a robust estimate of GPS positioning error that accounts for both time misalignment and systematic offsets.

\subsection{Analysis of Results}

Across all categories in Table~\ref{tab:five_by_three}, the system achieves sub-2m mean tracking error with bounded variability, indicating consistent performance under varied operating conditions. The effect of aerial coverage is clear in the linear trials: with three drones, the mean error is lowest at 0.94m (STD 0.74m), increasing to 1.04m (STD 0.85m) with two drones, and 1.11m (STD 0.85m) with a single drone. We can see the algorithm improves as more drones track the same marine robot, even though in the one-drone case, the marine robot is still accurately localized. Motion complexity degrades accuracy in a predictable but controlled manner. Non-linear trajectories tracked by three drones yield a mean error of 1.52m (STD 1.44 m), and the hardest regime with repeated sharp U-turns reaches 1.73 m (STD 1.48 m). These settings stress the image-space association and increase the likelihood of short detection dropouts or transient ID switches. Nonetheless, the hybrid matching that blends image IOU with GNSS-space proximity, together with EKF filtering, limits drift; and our translation-only ICP pre-alignment removes systematic inter-receiver offsets so that residuals primarily reflect instantaneous estimation uncertainty rather than bias.

\subsection{Sources of Error}
We identify several key sources of error that limit the accuracy of our results. \textbf{Estimation Uncertainty:} Although generally accurate, the vision-based tracking system faces difficulties in precise center localization, which introduces minor positional deviations. Additionally, the limited amount of data from the robot's field operations restricted our ability to fine-tune YOLO, hindering its performance. The model occasionally has extended detection failures, which severely disrupt trajectory tracking. \textbf{Sensor Noise:} The computation of our approach uses multiple variables that are prone to error. GNSS signals may be distorted by multipath interference or atmospheric delays; altitude estimates can drift with barometric pressure; and angular displacements are affected by lens distortion. Reliable computation further requires clock synchronization across different drones and the GNSS receiver, but timing imperfections add error. Environmental effects such as wind vibrations and temporary signal loss exacerbate these issues, jointly contributing to errors. Moreover, the accuracy of GNSS measurements and position estimation strongly impacts the performance of Algorithm~\ref{alg:align}. Inaccurate localization can cause ID misalignments across drones, which in turn hinder reliable aggregation of the estimations.

\begin{figure}[h]
  \centering
  \begin{subfigure}[t]{0.45\columnwidth}
    \centering
    \includegraphics[width=\linewidth]{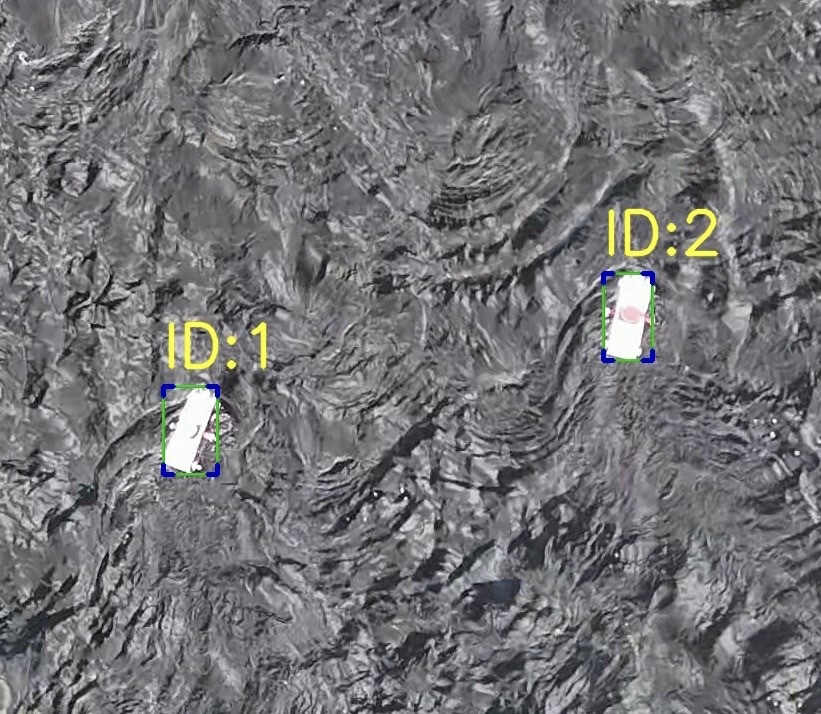}
    \caption{Hybrid Matching }
    \label{fig:left}
  \end{subfigure}\hfill
  \begin{subfigure}[t]{0.455\columnwidth}
    \centering
    \includegraphics[width=\linewidth]{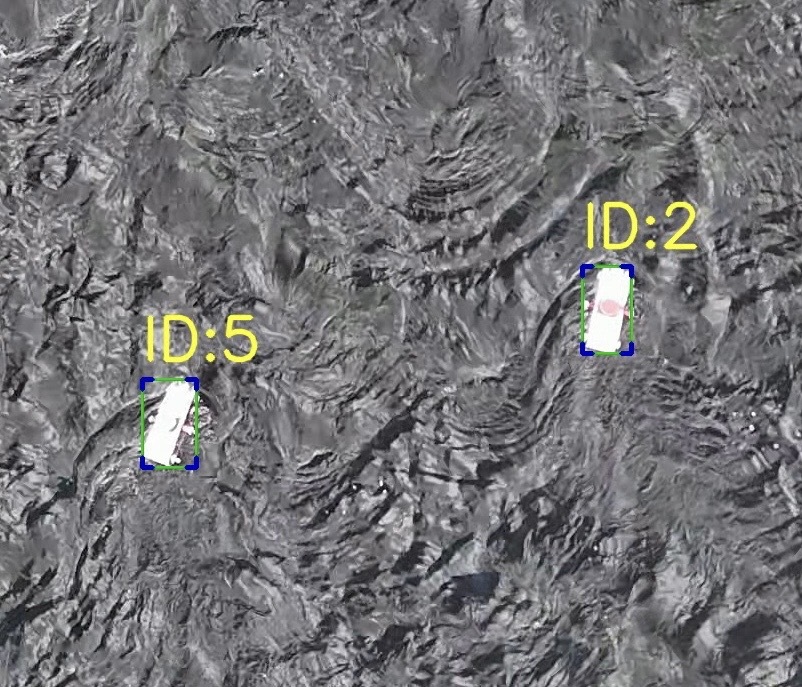}
    \caption{IOU matching}
    \label{fig:right}
  \end{subfigure}
  \caption{ID switches under strong air turbulence. Both images are from the same frame at the end of the same trial in Category 1 experiments. (a) Hybrid matching (IOU + normalized GNSS distance, see Section~\ref{subsec:hybrid}) maintains consistent IDs. (b) IOU-only matching suffers from severe ID switching.}
  \label{fig:idswitch}
\end{figure}

\subsection{Evaluating ID Stability Under Air Turbulence}
To demonstrate the stability of our hybrid matching, we evaluate the same tracking system but with different matching methods on the selected trials from Category, where it had many air turbulences, and then compare the results. Such turbulence caused rapid and severe camera shakes, often leading to errors in ID association during tracking. In these challenging trials, the IOU-based tracking system averages 1.33 ID switches per drone per 500 m, whereas our hybrid matching achieves zero switches. An example of an ID switch is shown in Fig.~\ref{fig:idswitch}.

\subsection{Hardware Testing}

To evaluate real-time performance, we deploy the tracking pipeline on an NVIDIA Jetson Xavier (32GB), a compute unit well-suited for embedded drone applications \cite{rojas2021board}. In the same trial segments used for the results with the presence of two marine robots, the Xavier processes the full pipeline in about 0.176 seconds per frame, averaging over 10 runs on more than 9,000 frames, yielding a 5 Hz position update rate. This frequency enables multiple marine robots to receive location updates from the drones in real time.

\section{Discussion and Conclusion}

In this paper, we propose and examine a tracking system that employs one or more drones to track marine robots operating near the water’s surface. The system integrates a vision-based detection module, a tracking algorithm, and a data filtering technique to achieve reliable performance. To ensure ID stability during long-range trials and under challenging conditions, we introduce a hybrid matching method that computes weighted similarity scores in both GNSS and image space. In addition, we design an inter-drone matching algorithm to align tracking IDs across different drones. Finally, we develop a data aggregation strategy that fuses complementary estimations based on confidence, further enhancing the overall accuracy and robustness of the system. 

The results presented in this paper indicate that multi-drone deployments strike an effective balance between accuracy and robustness for long, mission-scale operations. In contrast, single-drone setups remain suitable for applications where higher error tolerances are acceptable, such as surface inspections or diver guidance. The accuracy gains demonstrated with three drones are likely to improve further with larger fleets, though this hypothesis requires dedicated experimental validation. 

To fully demonstrate the algorithm’s utility, financial cost must be considered. A single onboard marine localization system costs roughly the same as four drones used in our experiments, yet it forces robots to remain on the surface and cannot localize multiple robots simultaneously. Our system runs in real time, transmitting high-frequency localization data to marine robots and enabling broader application potential.

In the future, we anticipate additional reductions in both mean error and variance through the use of more extensively trained YOLO object detection models and by upgrading the state estimation pipeline from an EKF to a particle filter, which would better capture the nonlinearities of the system. We also aim to develop a voting-based decentralized approach for ID alignment to further improve consistency. Furthermore, it would be valuable to evaluate the algorithm’s robustness in the face of more intense weather conditions and cases where, for certain frames, some or all the drones suffer data loss. Finally, future work could explore applying the proposed algorithm to moving objects in the water, as well as tracking groups of objects or marine creatures, such as schools of fish, thereby demonstrating the broader applicability of this approach to marine research and conservation.

% Not mention the refraction

\section{Acknowledgement}
The authors gratefully thank Chloe Si, Ali, Sasha Nicolas Dolgopolyy, Independent Robotics (IR), and the broader community of the McGill Mobile Robotics Lab (MRL) for their crucial assistance in field trials.

\printbibliography

\end{document}

%% file: drawing.tex
\tdplotsetmaincoords{70}{120}

{\scriptsize
\begin{tikzpicture}[tdplot_main_coords, scale=2]

  \def\R{2}
  \def\H{1}
  \def\angle{20}

  \pgfmathsetmacro{\X}{\R*cos(\angle)}
  \pgfmathsetmacro{\Y}{\R*sin(\angle)}
  \coordinate (O) at (0,0,0);

  \pgfmathsetmacro{\tanalpha}{\H/\R}

  % Define key points
  \coordinate (T) at (0, 0, \H);
  \coordinate (L) at (0.75*\R,\R,0); % robot1
  \coordinate (R) at (0,\R,0);

  % --- new points ---
  \coordinate (R2) at (0,\R,0);
  % \coordinate (R2) at (0.8, 2.4, -0.2); %  (-1.2,-1.1,-1.4);   % robot2, opposite side on base
  \coordinate (D3) at (-1.2, 0.8, 1.3); % drone2, slightly offset
  \coordinate (D2) at (-1.2, -2.2, 1.5); % drone3, another offset
  \coordinate (D2p) at (-1.2, -2.2, -0.75 *\H); % drone2, another offset
  \coordinate (D3p) at (-1.2, 0.8, -0.25 *\H); % drone3, another offset

  \pgfmathsetmacro{\ax}{0.25*\R/tan(\angle)}

  % center of the ellipse
  \coordinate (C) at (0, 0.25*\R, 0.75*\H); % center point for the square of drone1
  \coordinate (C2) at ($ (D2) + (-0.69, 0.12*\R, -0.7*\H) $); % center point for the square of drone2

  % THE PREVIOUS GOOD ONE
  % \coordinate (C3) at ($ (D3) + (0.0, 0.0*\R, -0.9*\H) $);

% COORDINATE IN LINE FROM D3 -- R2
\coordinate (C3) at ($(D3)!0.37!(R2)$);

  % corners (same offsets as for C, but centered at C2)
  \coordinate (C2a) at ($(-0.4,  0.1*\H,  0.1*\R)+(C2)$);
  \coordinate (C2b) at ($( 0.4,  0.1*\H,  0.1*\R)+(C2)$);
  \coordinate (C2c) at ($( 0.4, -0.1*\H, -0.1*\R)+(C2)$);
  \coordinate (C2d) at ($(-0.4, -0.1*\H, -0.1*\R)+(C2)$);

  % Dashed lines from tip T to the four corners of the plane
    \draw[dashed, gray] (D2) -- (C2a);
    \draw[dashed, gray] (D2) -- (C2b);
    \draw[dashed, gray] (D2) -- (C2c);
    \draw[dashed, gray] (D2) -- (C2d);

%%%%%%%%%%%%%%%%%%%%%%%%%%%%%%%%%%%%%%%%%%%%%%%%%%%%%%%%%%%%%
% --- rotate the drone-3 panel about its center C3 ---
\tdplotsetrotatedcoordsorigin{(C3)}
\tdplotsetrotatedcoords{-45}{-11.3}{45} %{-90}{-90}{90} % {yaw}{pitch}{roll}

\begin{scope}[tdplot_rotated_coords]
  % corners in the local (rotated) frame
  \coordinate (C3a) at (-0.4,  0.1*\H,  0.1*\R);
  \coordinate (C3b) at ( 0.4,  0.1*\H,  0.1*\R);
  \coordinate (C3c) at ( 0.4, -0.1*\H, -0.1*\R);
  \coordinate (C3d) at (-0.4, -0.1*\H, -0.1*\R);

  % draw the pane using the named corners
  \filldraw[fill=gray!45, draw=blue!70, thick, opacity=0.5]
    (C3a)--(C3b)--(C3c)--(C3d)--cycle;
  \draw[dashed, blue] ($(C3a)!0.5!(C3d)$) -- ($(C3b)!0.5!(C3c)$);
\end{scope}

  \draw[dashed, gray] (D3) -- (C3a);
  \draw[dashed, gray] (D3) -- (C3b);
  \draw[dashed, gray] (D3) -- (C3c);
  \draw[dashed, gray] (D3) -- (C3d);
  
  \coordinate (A) at (0.1875*\R, 0.25*\R, 0.75 *\H);
  % \coordinate (B) at (0.1875*\R, 0.25*\R, -1.3);

  % Circle base
  \draw[dashed] (0:\R) arc (0:360:\R);

  % Slanted cone sides
  \draw[thick] (T) -- (L);       % slanted edge at angle
  \draw[thick] (T) -- (R);  % vertical slanted edge

  \draw[thick, gray] (D2) -- (L);
  \draw[thick, gray] (D2) -- (R2);

  \draw[thick, darkgray] (D3) -- (L);
  \draw[thick, darkgray] (D3) -- (R2);

  % blue opac plane
  % for drone 1
  \filldraw[fill=gray!45, draw=blue!70, thick, opacity=0.5]
       ($ (-0.4, 0.1*\H, 0.1*\R)+ (C)$) -- ($ (0.4, 0.1*\H, 0.1*\R)+ (C)$) -- ($ (0.4, -0.1*\H, -0.1*\R)+ (C)$) -- ($ (-0.4, -0.1*\H, -0.1*\R)+ (C)$)  -- cycle;
  \draw[dashed, blue] ($ (-0.4,0, 0)+ (C)$) -- ($ (0.4,0, 0)+ (C)$);

  % for drone 2
  \filldraw[fill=gray!45, draw=blue!70, thick, opacity=0.5]
        (C2a) -- (C2b) -- (C2c) -- (C2d) -- cycle;
  \draw[dashed, blue] ($(C2a)!0.5!(C2d)$) -- ($(C2b)!0.5!(C2c)$);

  % for drone 3
  % \filldraw[fill=gray!45, draw=blue!70, thick, opacity=0.5]
  %       (C3a) -- (C3b) -- (C3c) -- (C3d) -- cycle;
  % \draw[dashed, blue] ($(C3a)!0.5!(C3b)$) -- ($(C3c)!0.5!(C3d)$);

  % tiny red dots (center + one interior point)
  \filldraw[green] (C) circle (0.4pt);
  \filldraw[red] (A) circle (0.4pt);
  % \filldraw[blue] (C3) circle (0.4pt);

  \coordinate (R2_proj_D2) at ($ (D2) + (-0.75, 0.12*\R, -0.7*\H) $);
  \filldraw[green] (R2_proj_D2) circle (0.4pt);

  \coordinate (A2) at ($ (C2) + (0.19, -0.02*\H, -0.01*\R) $); 
  \filldraw[red] (A2) circle (0.4pt);  

  \coordinate (R1_proj_D3) at at ($(D3)!0.38!(L)$);
  \filldraw[red] (R1_proj_D3) circle (0.4pt);

  \coordinate (R2_proj_D3) at ($(D3)!0.37!(R2)$);

  \filldraw[green] (R2_proj_D3) circle (0.4pt);
  
  % \coordinate (A3) at ($ (C3) + (0.12, 0.02*\H, 0.02*\R) $);
  % \coordinate (A3) at ($ (C3) + (0, 0, 0.18*\H) $);
  % \coordinate (A3) at ($(C3a)!0.5!(C3b)$);
  % \filldraw[red] (A3) circle (0.4pt);

  % Dashed lines from tip T to the four corners of the plane
    \draw[dashed, gray] (T) -- ($ (-0.4,  0.1*\H,  0.1*\R) + (C) $);
    \draw[dashed, gray] (T) -- ($ ( 0.4,  0.1*\H,  0.1*\R) + (C) $);
    \draw[dashed, gray] (T) -- ($ ( 0.4, -0.1*\H, -0.1*\R) + (C) $);
    \draw[dashed, gray] (T) -- ($ (-0.4, -0.1*\H, -0.1*\R) + (C) $);

  % Dashed lines  
  \draw[dashed] (D2p) -- (D2);   
  \draw[dashed] (D3p) -- (D3);   
  \draw[dashed] (D2p) -- (R2);
  \draw[dashed] (D2p) -- (L);
  \draw[dashed] (D3p) -- (R2);
  \draw[dashed] (D3p) -- (L);

  % Central axis
  \draw[dashed] (0,0,0) -- (T) node[pos=0.35, left] {$A$};

  % Base edges
  \draw[dashed] (0, 0, 0) -- (L) node[midway, below left, xshift=2pt, yshift=1pt] {$D_r$};
  \draw[dashed] (0, 0, 0) -- (R) node[pos=0.35, above right, xshift=2pt, yshift=1pt] {$D$};

  \draw[thick] (L) -- (R) node[midway, below right] {$D_x$};

  % Optional: highlight apex and base
  \filldraw[violet] (T) circle (1pt) node[above=4pt, text=blue] {\small \textbf{Drone1}};
  \filldraw[red] (L) circle (1pt) node[below=4pt, text=blue] {\small \textbf{Robot1}};
  \filldraw[green] (R2) circle (1pt) node[below=4pt, text=blue] {\small \textbf{Robot2}};
  \filldraw[violet] (D2) circle (1pt) node[above right, text=blue] {\small \textbf{Drone2}};
  \filldraw[violet] (D3) circle (1pt) node[above left, text=blue] {\small \textbf{Drone3}};

% --- helper midpoints that define each panel’s dashed in-plane axis
\coordinate (D1eL) at ($ (-0.4, 0, 0) + (C)$);
\coordinate (D1eR) at ($ ( 0.4, 0, 0) + (C)$);

\coordinate (D2mL) at ($(C2a)!0.5!(C2d)$);
\coordinate (D2mR) at ($(C2b)!0.5!(C2c)$);

\coordinate (D3mL) at ($(C3a)!0.5!(C3d)$);
\coordinate (D3mR) at ($(C3b)!0.5!(C3c)$);

% === Drone1: right angle at green point C ===
\coordinate (Cnorm)  at ($(C)!-0.18!(R2)$);              % along drone1→Robot2 line
\coordinate (Ctang) at ($ (C) + 0.18*(D1eR) - 0.18*(D1eL) $);  % parallel to dashed in-plane axis
\pic[draw,thick,angle radius=5pt] {right angle = Cnorm--C--Ctang};

% === Drone2: right angle at green point R2_proj_D2 ===
\coordinate (P2)     at (R2_proj_D2);
\coordinate (P2norm) at ($(R2_proj_D2)!-0.18!(R2)$);             % along drone2→Robot2 line
\coordinate (P2tang) at ($ (P2) + 0.18*(D2mR) - 0.18*(D2mL) $); % along panel dashed axis
\pic[draw,thick,angle radius=5pt] {right angle = P2norm--P2--P2tang};

% === Drone3: right angle at green point R2_proj_D3 ===
\coordinate (P3)     at (R2_proj_D3);
\coordinate (P3norm) at ($(R2_proj_D3)!-0.18!(R2)$);             % along drone3→Robot2 line
\coordinate (P3tang) at ($ (P3) + 0.18*(D3mR) - 0.18*(D3mL) $); % along panel dashed axis
\pic[draw,thick,angle radius=5pt] {right angle = P3norm--P3--P3tang};

% Right angle between axis A (O--T) and line O--R2, oriented the other way
\coordinate (Ao)  at ($(O)!0.18!(T)$);   % along A toward T
\coordinate (R2o) at ($(O)!0.18!(R2)$);  % toward Robot2
\pic[draw,thick,angle radius=7pt] {right angle = Ao--O--R2o};

% Right angle at Robot 2
\coordinate (Rneg) at (0,0.9*\R,0);
\coordinate (Rpos) at (0.1*\R,\R,0);
\pic[draw,thick,angle radius=7pt] {right angle = Rneg--R--Rpos};

  \pic [draw, angle radius=15pt, angle eccentricity=1.3, "$\scriptstyle \theta_{x}$"] {angle = L--T--R};
  
  \path (0,0,0) -- (L) coordinate[pos=0.2] (LO);
  \path (0,0,0) -- (R) coordinate[pos=0.2] (RO);

  \path (0, 0, 0) -- (-2, 1, 0) coordinate[pos=0.2] (NO);

  \pic [draw, angle radius=20pt, angle eccentricity=1.4, "\scriptsize $\beta$"] {angle = LO--O--RO};

  \draw[->, dashed, brown](0, 0, 0) -- (-2, 1, 0) node [pos=0.8, left] {N};

  \pic [draw, brown, angle radius=15pt, angle eccentricity=1.5, "\scriptsize $\psi$"] {angle = RO--O--NO};

\end{tikzpicture}
}